\documentclass[letterpaper,12pt]{article} 
\usepackage{amsmath} 
\pdfoutput=1 
\usepackage{graphicx}
\usepackage{array}
\usepackage{tabu}
\usepackage{mathrsfs}

\usepackage{framed, color}
\usepackage{mdframed, color}

    \usepackage[utf8]{inputenc}
    \usepackage[T1]{fontenc}
    \usepackage{lmodern}
    \usepackage[x11names]{xcolor}
    \usepackage{framed}
    \colorlet{shadecolor-pink}{LavenderBlush2}
    \colorlet{framecolor}{Red1}
    \usepackage{lipsum}

    {\endMakeFramed}

    \newenvironment{frshaded*}{%
    \MakeFramed {\advance\hsize-\width \FrameRestore}}%
    {\endMakeFramed}

\usepackage{tcolorbox}

\usepackage{amsthm,color}
\theoremstyle{definition}

\mdfdefinestyle{example1}{%
    linecolor=blue,
    outerlinewidth=2pt,
    bottomline=false,
    leftline=false,rightline=false,
    skipabove=\baselineskip,
    skipbelow=\baselineskip,
    frametitle=\mbox{},
}

\mdfdefinestyle{exampledefault}{%
rightline=true,
leftline = true,
innerleftmargin=10,innerrightmargin=10,
frametitlerule=true,frametitlerulecolor=green,
frametitlebackgroundcolor=yellow,
frametitlerulewidth=2pt}

\usepackage{url}            
\usepackage{booktabs}       
\usepackage{amsfonts}       
\usepackage{nicefrac}       
\usepackage{microtype}      

\definecolor {shadecolor}{rgb}{1, 0.8, 0.3}
\definecolor {eqn-descrpt-aqua}{RGB}{210, 240, 235}
\definecolor {eqn-descrpt-frame-aqua}{RGB}{83, 111, 121}
\definecolor {med-aqua}{RGB}{165, 210, 200}
\definecolor {eqn-celadon}{RGB}{215, 240, 221}
\definecolor {eqn-frame-celadon}{RGB}{33, 130, 125}

\title{\textbf{Free Energy Minimization Using the 2-D Cluster Variation Method: Initial Code Verification and Validation} \\
\vspace{10 mm} } 
\date{Revision Date: 2019-06-25\\
  Version 2}

\author{Alianna J. Maren \\
  Northwestern University School of Professional Studies\\
  Master of Science in Data Science Program\\  
  and\\
  Themasis  \\
  Themasis Technical Report TR-2018-01v2\\ 
  {\tt alianna.maren@northwestern.edu}\\
  {\tt alianna@aliannajmaren.com} 
  }

\begin{document}

\maketitle

\newpage


\abstract{A new approach for artificial general intelligence (AGI), building on neural network deep learning architectures, can make use of one or more hidden layers that have the ability to continuously reach a free energy minimum even after input stimulus is removed, allowing for a variety of possible behaviors. One reason that this approach has not been developed until now has been the lack of a suitable free energy equation; one that would avoid some of the difficulties known in Hopfield-style neural networks. The cluster variation method (CVM) offers a means for characterizing 2-D local pattern distributions, or \textit{configuration variables}, and provides a free energy formalism in terms of these configuration variables. The equilibrium distribution of these configuration variables is defined in terms of a single interaction enthalpy parameter, \textit{h}, for the case of equiprobable distribution of bistate (neural/neural ensemble) units. For non-equiprobable distributions, the equilibrium distribution can be characterized by providing a fixed value for the fraction of units in the active state ($x_1$), corresponding to the influence of a per-unit activation enthalpy, together with the pairwise interaction enthalpy parameter \textit{h}. 

This paper provides verification and validation (V\&V) for code that computes the configuration variable and thermodynamic values for 2-D CVM grids characterized by different interaction enthalpy parameters, or \textit{h}-values. This means that there is now a working foundation for experimenting with a 2-D CVM-based hidden layer that can, as an alternative to responding strictly to inputs, also now independently come to its own free energy minimum. Such a system can also return to a free energy-minimized state after it has been perturbed, which will enable a range of input-independent behaviors that have not been hitherto available. A further use of this 2-D CVM grid is that by characterizing different kinds of patterns in terms of their corresponding \textit{h}-values (together with their respective fraction of active-state units), we have a means for quantitatively characterizing different kinds of neural topographies. This further allows us to connect topographic descriptions (in terms of local patterns) with free energy minimization, allowing a first-principles approach to characterizing topographies and building new computational engines. }

\vspace{10pt}

\textbf{Keywords:} artificial intelligence; neural networks; deep learning; statistical thermodynamics; free energy; free energy minimization; cluster variation method; entropy; brain networks; neural connectivity

\pagebreak

\section{Introduction and Overview}
\label{sec:intro-and-overview}
%

This article documents the verification and validation (V\&V) results for the first two stages of code development for free energy minimization within a 2-D cluster variation method (CVM) system. 

The intention is that this 2-D CVM system can have its free energy minimized independent of its use in any other process. Ultimately, the 2-D CVM system will be inserted as a single layer into a neural network, creating a new form of computational engine, which I call the CORTECON, standing for a COntent-Retentive, TEmporally-CONnected neural network, first described in \cite{Maren_1993_Free-energy-as-driving-function} and \cite{Maren-Schwartz-Seyfried_1992_Config-entropy-stabilizes}, both of which presented early results using a 1-D CVM.  

\begin{figure}
\label{fig:computational-engine-1}
  \centering
  \fbox{
  \rule[0cm]{0cm}{0cm}\rule[0cm]{0cm}{0cm}	
  \includegraphics [trim=0.0cm 0cm 0.0cm 0cm, clip=true,   width=0.9\linewidth]{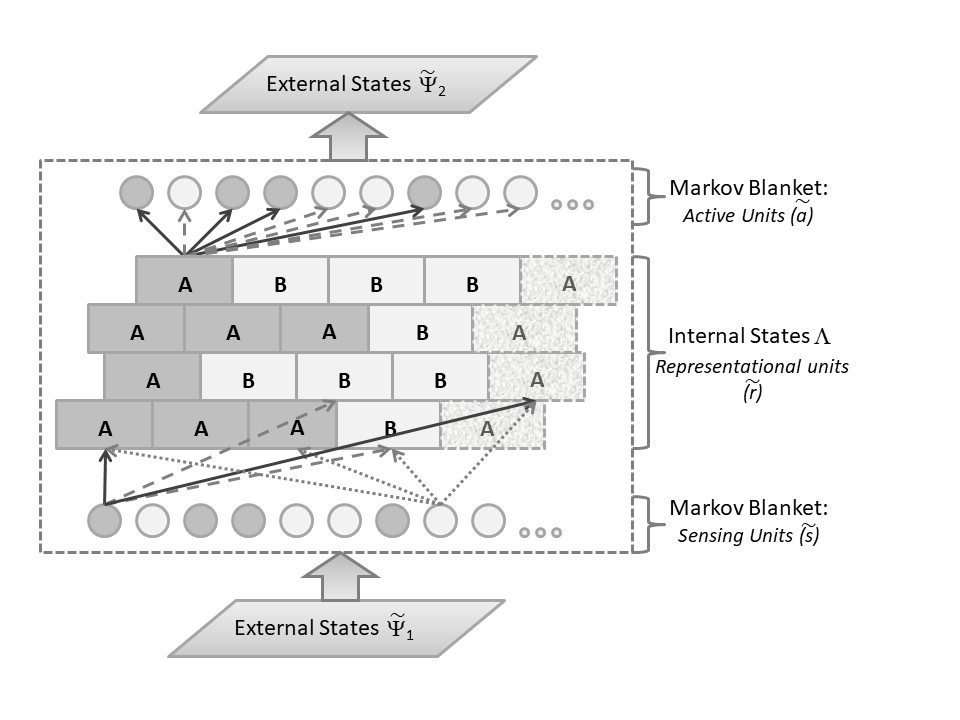}} 
  \vspace{3mm} 
  \caption{Illustration of a cluster variation method (CVM)-based computational engine, for which the Markov blanket of sensing and active units corresponds to input and output layers (see Friston \cite{Friston-et-al_2015_Knowing-ones-place-free-energy-pattern-recognition}). Unique to the approach advanced here, the computational layer is composed as a 2-D CVM, for which the free energy equation can be explicitly written, and the free energy minimum can be found either analytically or computationally, depending on the parameters used. The CVM layer comprises  the internal or representational units $\tilde{r}$ , and cannot communicate with the external field (shown in two parts for visualization purposes only). However, units within the representational layer can receive inputs from the sensory units $\tilde{s}$ and send signals to the active $\tilde{a}$ units. The sensory units can receive inputs from external stimulus, and send signals to the representational units. The active units can receive inputs from the representational units, and send signals to the external output units. (In the notional view advanced by Friston (op. cit.), a broader set of interactions is allowed; for simplicity in this engine, the interaction pathways have been streamlined.)}
\label{fig:CVM-2-D_comput-engine_crppd_2017-05-13}
\end{figure}

This work described here focuses on a 2-D CVM grid, which can operate as both a hidden layer and as a independent functional unit, with the ability to achieve a free energy minimum when there is no extraneous signal coming into the layer, is shown in Figure~\ref{fig:computational-engine-1}. 

This notion of using a 2-D CVM as a \textit{computational engine's} hidden layer advances ideas originally proposed in \cite{Maren_1993_Free-energy-as-driving-function} and \cite{Maren-Schwartz-Seyfried_1992_Config-entropy-stabilizes}, along with \cite{Maren_2016_CVM-primer-neurosci}, and further incorporates (and makes practical) ideas put forth by Karl Friston, whose notation was adopted for Figure~\ref{fig:computational-engine-1}. This figure illustrates the \textit{computational engine} using Friston's notion of a set of computational (representational) units separated from an external system by a Markov blanket. It also allows for the variational Bayes (free energy minimization) approach described by Friston \cite{Friston-et-al_2015_Knowing-ones-place-free-energy-pattern-recognition}, \cite{Friston_2010_Free-energy-principle-unified-brain-theory}, and \cite{Friston_2013_Life-as-we-know-it}. (For a detailed derivation, see Maren \textbf{ (xxx)}.)

In brief, Friston (building on work by Beal \cite{Beal_2003_Variational-algorithm-approx-Bayes-inference}) proposes a computational system in which a Markov blanket separates the computational (representational) elements of the engine from external events, as shown in Figure~\ref{fig:CVM-2-D_comput-engine_crppd_2017-05-13}. The communication between the external system elements (denoted $\tilde{\psi}$) with those of the representational system (denoted $\lambda$ or $\tilde{r}$) are mediated by two distinct layers or components of the Markov blanket; the sensing ($\tilde{s}$) elements and the action ($\tilde{a}$) ones. 

\vspace{3mm} 

This article provides V\&V for the first two code development stages: 

\begin{enumerate} \itemsep0pt 
\item \textbf{Computing values for the configuration variables in a 2-D system} -- for various values of the interaction enthalpy parameter \textit{h}, and 
\item \textbf{Computing the thermodynamic quantities associated with the 2-D system} -- given the set of configuration variables, it is possible to then compute enthalpy, entropy, and free energy. 
\end{enumerate}

Most crucially, the code incorporates a free energy minimization process, so that once an initial (randomly-generated pattern) has been created, it is adjusted in a two-stage process: 

\begin{enumerate} \itemsep0pt 
\item \textbf{Achieve desired $x_1$ specification} -- this allows us to implicitly enfold a nominal per-unit activation energy (where the relationship between this parameter $\varepsilon_0$ and $x_1$ cannot be explicitly stated at this time), and 
\item \textbf{Achieve free energy minimization for the given set of $x_1$, $h$-values} -- typically, the 2-D CVM grid needs to have state changes in its various units to achieve a free energy minimum. 
\end{enumerate}

\section{The Configuration Variables}
\label{sec:config-variables}
%

The first V\&V aspect of the task documented here was to ensure that the configuration variables for the 2-D grid were counted correctly.

\begin{enumerate} \itemsep0pt 
\item \textbf{\textit{Configuration variable definitions}} -- including how they are counted in the 2-D CVM grid, 
\item \textbf{\textit{2-D CVM grid specifications}} -- size and wrap-arounds, and 
\item \textbf{\textit{V\&V results}} -- configuration variable counts for select examples.
\end{enumerate}

%
\subsection{Introducing the Configuration Variables}
\label{subsec:Introducing-config-vars}
%

The cluster variation method, introduced by Kikuchi \cite{Kikuchi_1951_Theory-coop-phenomena} and refined by Kikuchi and Brush \cite{Kikuchi-Brush_1967_Improv-CVM}, uses an entropy term that includes not only the distribution into simple ``on'' and ``off'' states, but also distribution into local patterns, or \textit{configurations}, as illustrated in the following figures.

A 2-D CVM is characterized by a set of \textit{configuration variables}, which collectively represent single unit, pairwise combination, and triplet values. The configuration variables are denoted as: 

\begin{itemize}
\setlength{\itemsep}{1pt}
\item $x_i$ - Single units, 
\item $y_i$ - Nearest-neighbor pairs, 
\item $w_i$ - Next-nearest-neighbor pairs, and 
\item $z_i$ - Triplets. 
\end{itemize}

\begin{figure} [ht]
    \centering
        \includegraphics[trim=4cm 7.7cm 5.5cm 7.7cm, clip=true,  width=0.9\textwidth]{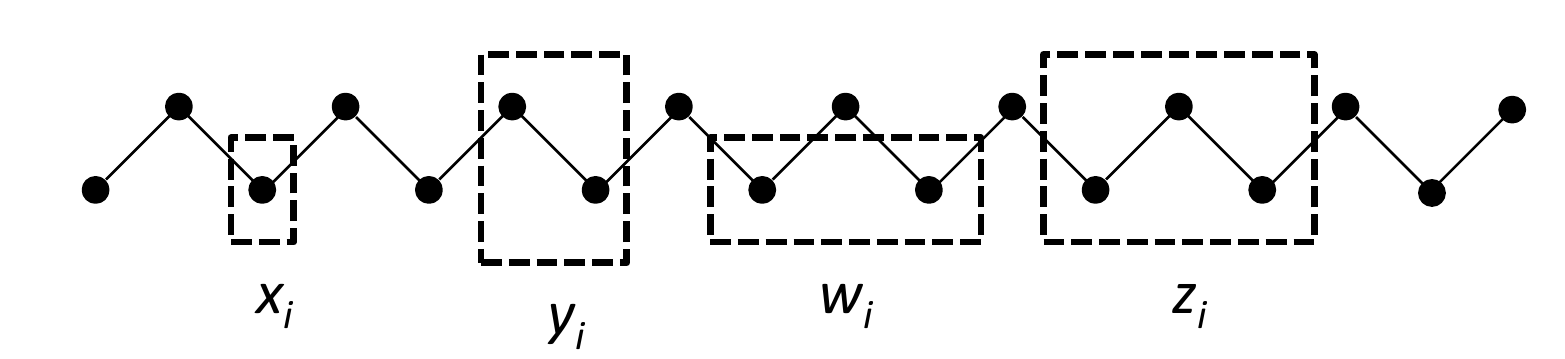}
    \caption{The 1-D single zigzag chain is created by arranging two staggered sets of $M$ units each. The configuration variables shown are $x_i$ (single units), $y_i$ (nearest-neighbors), $w_i$ (next-nearest-neighbors), and $z_i$ (triplets).}
    \label{fig:single_zigzag_chain_lbld}
\end{figure}

These configuration variables are illustrated for a single zigzag chain in Figure~\ref{fig:single_zigzag_chain_lbld}.
 
\begin{figure}[ht]
    \centering
        \includegraphics[trim=0cm 0cm 0cm 0cm, clip=true,  width=0.6\textheight]{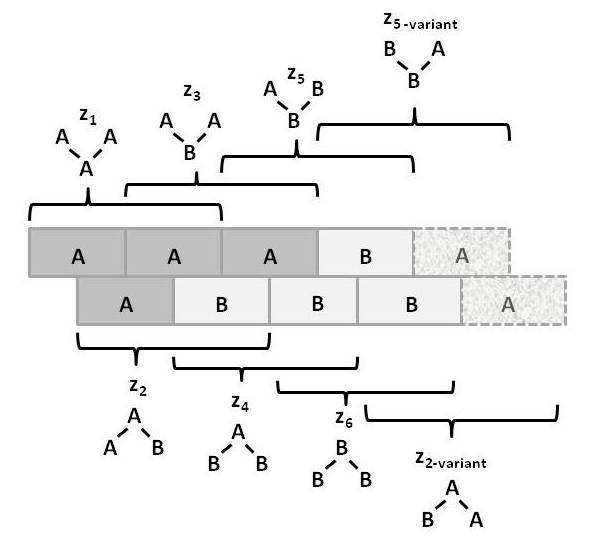}
    \caption{The six ways in which the configurations $z_i$ can be constructed.}
    \label{fig:CVM-1-D_base-graph2}
\end{figure}

For a bistate system (one in which the units can be in either state \textbf{A} or state \textbf{B}), there are six different ways in which the triplet configuration variables ($z_i$) can be constructed, as shown in Figure~\ref{fig:CVM-1-D_base-graph2}, and also in  Table~\ref{tbl:config-variables-table}.

Notice that within Figure~\ref{fig:CVM-1-D_base-graph2}, the triplets $z_2$ and $z_5$ have two possible configurations each: \textbf{A}-\textbf{A}-\textbf{B} and \textbf{B}-\textbf{A}-\textbf{A} for $z_2$, and \textbf{B}-\textbf{B}-\textbf{A} and \textbf{A}-\textbf{B}-\textbf{B} for $z_5$. This means that \textbf{\textit{there is a degeneracy factor of 2}} for each of the $z_2$ and $z_5$ triplets.

\begin{figure}[ht]
  \centering
  \fbox{
  \rule[-.5cm]{0cm}{4cm}\rule[-.5cm]{0cm}{0cm}	
  \includegraphics [trim=0.0cm 0cm 0.0cm 0cm, clip=true,   width=0.95\linewidth]{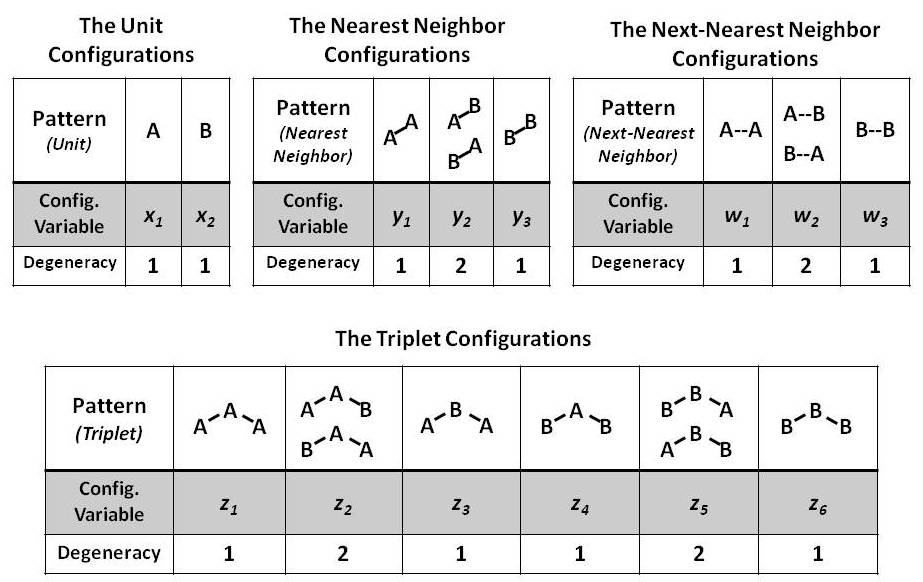}} 
  \vspace{3mm} 
  \caption{Illustration of the \textit{configuration variables} for the cluster variation method, showing the ways in which the configuration variables $y_i$, $w_i$, and $z_i$ can be constructed, together with their degeneracy factors $\beta_i$ and $\gamma_i$.}   
\label{fig:Config-var-weights_v3_crppd_2017-05-17}
\end{figure}

The degeneracy factors $\beta_i$ and $\gamma_i$ (number of ways of constructing a given configuration variable) are shown in Figure~\ref{fig:Config-var-weights_v3_crppd_2017-05-17}; $\beta_2 = 2$,  as $y_2$ and $w_2$ can be constructed as either \textbf{A}-\textbf{B} or as \textbf{B}-\textbf{A} for $y_2$, or as \textbf{B}-~-\textbf{A} or as \textbf{A}-~-\textbf{B} for $w_2$. Similarly, $\gamma_2 = \gamma_5 = 2$ (for the triplets), as there are two ways each for constructing the triplets $z_2$ and $z_5$. All other degeneracy factors are set to 1.

\vspace{3mm} 

\begin{table}[t]
  \caption{Configuration Variables for the Cluster Variation Method}
  \label{tbl:config-variables-table}
  \centering
  \begin{tabular}{lcc}
    \toprule
    Name     & Variable     & Instances \\
    \midrule
    Unit & $x_i$  & 2     \\
    Nearest-neighbor     & $y_i$ & 3      \\
    Next-nearest-neighbor     & $w_i$       & 3  \\
    Triplet     & $z_i$       & 6  \\    
    \bottomrule
  \end{tabular}
\end{table}

%
\subsection{Counting the Configuration Variables}
\label{subsec:counting-config-vars}
%

\begin{figure}[ht]
  \centering
  \fbox{
  \rule[-.5cm]{0cm}{4cm}\rule[-.5cm]{0cm}{0cm}	
  \includegraphics [trim=0.0cm 0cm 0.0cm 0cm, clip=true,   width=0.95\linewidth]{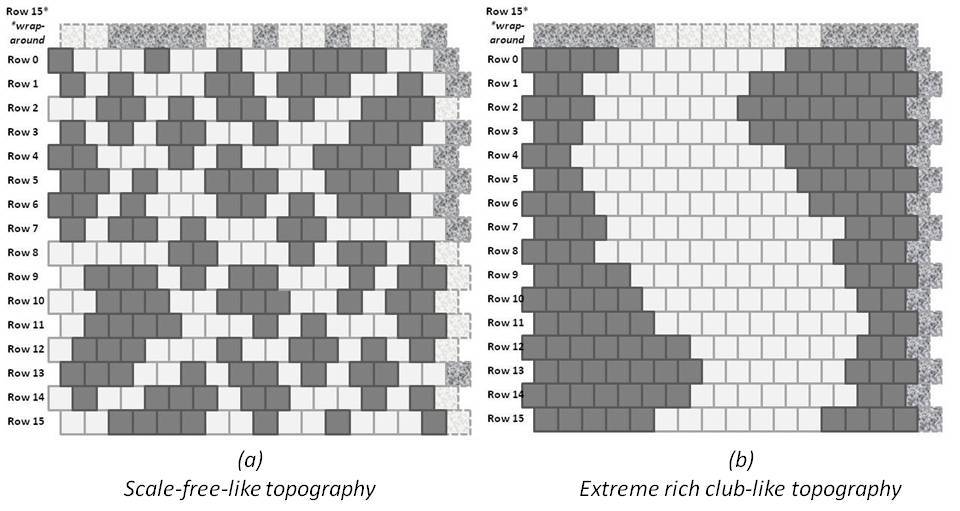}} 
  \vspace{3mm} 
  \caption{Illustration of the two different grids for experiments with the 2-D CVM system.}   
\label{fig:CVM-2-D_rich-club_and_scale-free_256-nodes}
\end{figure}

To experiment with the 2-D CVM system, I constructed various grids of 256 (16 x 16) units each, as illustrated in Figure~\ref{fig:CVM-2-D_rich-club_and_scale-free_256-nodes}. 

I decided to use a 16 x 16 grid for several reasons:

\begin{enumerate} \itemsep0pt 
\item \textbf{\textit{Sufficient variety in local patterns}} -- I was able to construct grids that illustrated several distinct kinds of topographies (each corresponding to different \textit{h}-values), 
\item \textbf{\textit{Sufficient nodes}} -- so that triplet-configuration extrema could be explored in some detail, and 
\item \textbf{\textit{Countability}} -- I needed to be able to manually count all the configuration values for a given 2-D grid configuration, and match them against the results from the program, as a crucial V\&V step.
\end{enumerate} 

One final advantage of the 16 x 16 grid layout was that the different grid configurations were both large enough to show diversity, but small enough so that I could create a figure illustrating the activation states (\textbf{A} or \textbf{B}) of each node, thus illustrating the detailed particulars of each configuration design. 

I began with manually-designed grid configurations, such as the two shown in Figure~\ref{fig:CVM-2-D_rich-club_and_scale-free_256-nodes}. These two configurations correspond (somewhat) to the notions of ``scale-free'' and ``rich-club'' topographies, as observed in various neural communities. (For references, please consult \cite{Maren_2016_CVM-primer-neurosci}.)

\vspace{3mm} 

These two different grid configurations are early attempts to characterize how the \textit{h}-values can be identified for grids with different total configuration variable values. The following Section~\ref{sec:v-and-v-thermodynamic-variables} will discuss \textit{h}-values in the context of the free energy equation. 

Both of these systems were created with the constraint of equiprobable occurrence of units in states (\textbf{A} or \textbf{B}; that is, $x_1 = x_2= 0.5$. This was done to facilitate the next V\&V step, which will be discussed in Section~\ref{sec:v-and-v-thermodynamic-variables}. Thus, for the configurations shown in Figure~\ref{fig:CVM-2-D_rich-club_and_scale-free_256-nodes}, both the (\textbf{\textit{a}}) and (\textbf{\textit{b}}) grids have 128 nodes each of units in state \textbf{A} and in state \textbf{B}.

The configuration on the left of Figure~\ref{fig:CVM-2-D_rich-club_and_scale-free_256-nodes} is an effort to build a ``scale-free-like'' system. The notion of a ``scale-free'' system is that the same kind of pattern replicates itself throughout various scales of observation in a system. Thus, for the ``scale-free-like'' configuration shown in Figure~\ref{fig:CVM-2-D_rich-club_and_scale-free_256-nodes} (\textbf{\textit{a}}), I created a design that was originally intended to be 180-degree symmetrical around a central axis (dihedral group-2 symmetry). Specifically, the left and right sides were to be identical in a rotated-180-degree sense. 

For ease in design of the ``scale-free-like'' system, I focused on creating a pattern on one side and duplicating it on the other. I used a paisley-like base pattern in order to create greater dispersion of values across the $z_i$ triplets; that is, I wanted to minimize having tightly-clustered islands that would yield little in the way of \textbf{A}-\textbf{B}-\textbf{A} and \textbf{B}-\textbf{A}-\textbf{B} triplets ($z_2$ and $z_5$, respectively). 

The practical limitation of attempting to fit various ``islands'' of \textbf{A} nodes (black) into a surrounding ``sea'' of \textbf{B} nodes (white) meant that there were not quite enough \textbf{B} nodes to act as borders around the more compact sets of \textbf{A} nodes. Thus, the pattern in the right half of grid (\textbf{\textit{a}}) is a bit more compressed than originally planned. 

\begin{figure}[ht]
  \centering
  \fbox{
  \rule[-.5cm]{0cm}{4cm}\rule[-.5cm]{0cm}{0cm}	
  \includegraphics [trim=0.0cm 0cm 0.0cm 0cm, clip=true,   width=0.95\linewidth]{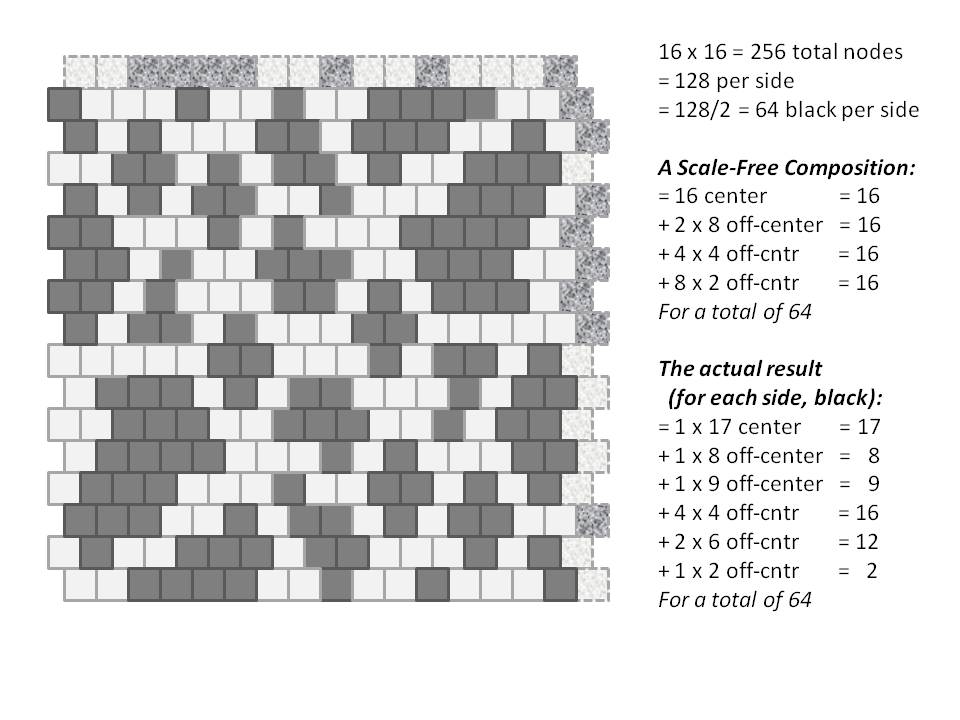}} 
  \vspace{3mm} 
  \caption{A 2-D CVM ``scale-free-like'' system with an equal number state \textbf{A} and state \textbf{B} nodes (128 nodes each).}   
\label{fig:CVM-2D_Scale-free_128-nodes_pg0_2016-10-26}
\end{figure}

The original plan was that out of 256 nodes in the grid, half (of the designed pattern) would be on the right, and half on the left; 128 nodes on each side. Of these, for each side, 64 nodes were to be in state \textbf{A} (black). Of these nodes (per side), sixteen (16 nodes) would be used to create a large, paisley-shaped island. The remaining 64 - 16 = 48 nodes would be used for smaller-sized islands; two islands of eight nodes each, etc. The plan is shown in Figure~\ref{fig:CVM-2D_Scale-free_128-nodes_pg0_2016-10-26}. The notation of ``center'' and ``off-center`` refers to the placement of the various islands; the largest (16-node) islands were to be placed more-or-less in the center of each of their respective (left or right) sides of the grid, and the remaining islands were to be ``off-center''; situated around their primary respective large islands.

The resulting patterns were close to the original plan, although not exactly the same. (Again, for details, see Figure~\ref{fig:CVM-2D_Scale-free_128-nodes_pg0_2016-10-26}.)

\vspace{3mm} 

Even though some changes had to be made to the original design plan, the original constraint, that the number of units in states \textbf{A} and \textbf{B} would be identical (128 nodes in each), was kept. The details are shown in Figure~\ref{fig:CVM-2D_Scale-free_128-nodes_pg0_2016-10-26}.

\begin{figure}[ht]
  \centering
  \fbox{
  \rule[-.5cm]{0cm}{4cm}\rule[-.5cm]{0cm}{0cm}	
  \includegraphics [trim=0.0cm 0cm 0.0cm 0cm, clip=true,   width=0.95\linewidth]{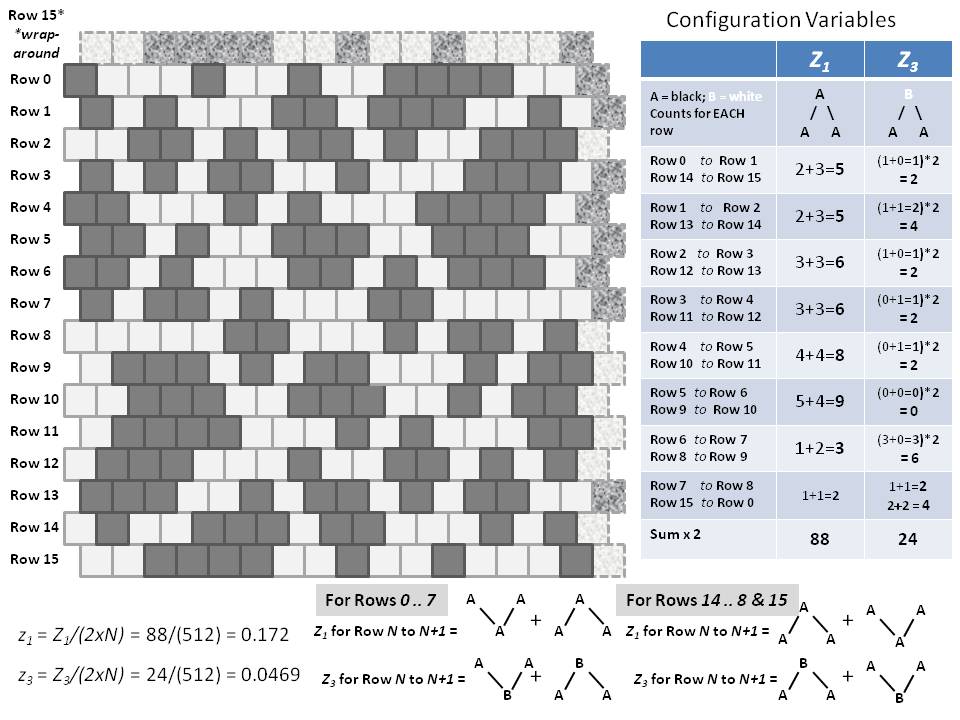}} 
  \vspace{3mm} 
  \caption{A 2-D CVM ``scale-free-like'' system with an equal number state \textbf{A} and state \textbf{B} nodes (128 nodes each).}   
\label{fig:CVM-2D_scale-free_2017-12-12}
\end{figure}
\vspace{3mm} 

The validation step for this stage of code development was to manually count all the configuration variables for several different configuration grids, such as the ones shown in Figure~\ref{fig:CVM-2-D_rich-club_and_scale-free_256-nodes}.

The counts for the ``scale-free-like'' grid shown in Figure~\ref{fig:CVM-2D_Scale-free_128-nodes_pg0_2016-10-26} are shown in Figure~\ref{fig:CVM-2D_scale-free_2017-12-12}. It suffices to say that the results from the manual counting (of all configuration variables) and those created by the computer code were identical. These held true across several different grids with different node configurations. 

\textbf{\textit{Note:}} To achieve the fractional variables shown in Figure~\ref{fig:CVM-1-D_base-graph2}, and also in  Table~\ref{tbl:config-variables-table}, the following relations are used: 

\begin{itemize}
\setlength{\itemsep}{1pt}
\item $x_i = X_i$, 
\item $y_i = Y_i/2$, for $i = 1, 3$ and $y_2 = Y_2/4$, accounting for the degeneracy with which $y_2$ occurs,
\item $w_i = W_i/2$, for $i = 1, 3$ and $w_2 = W_2/4$, accounting for the degeneracy with which $w_2$ occurs, and 
\item $z_i = Z_i/2$, for $i = 1, 3, 4, 6$ and $z_2 = Z_2/4$, $z_5 = Z_5/4$, accounting for the degeneracy with which $z_2$ and $z_5$ occur. 
\end{itemize}
 
\begin{figure}[ht]
  \centering
  \fbox{
  \rule[-.5cm]{0cm}{4cm}\rule[-.5cm]{0cm}{0cm}	
  \includegraphics [trim=0.0cm 0cm 0.0cm 0cm, clip=true,   width=0.95\linewidth]{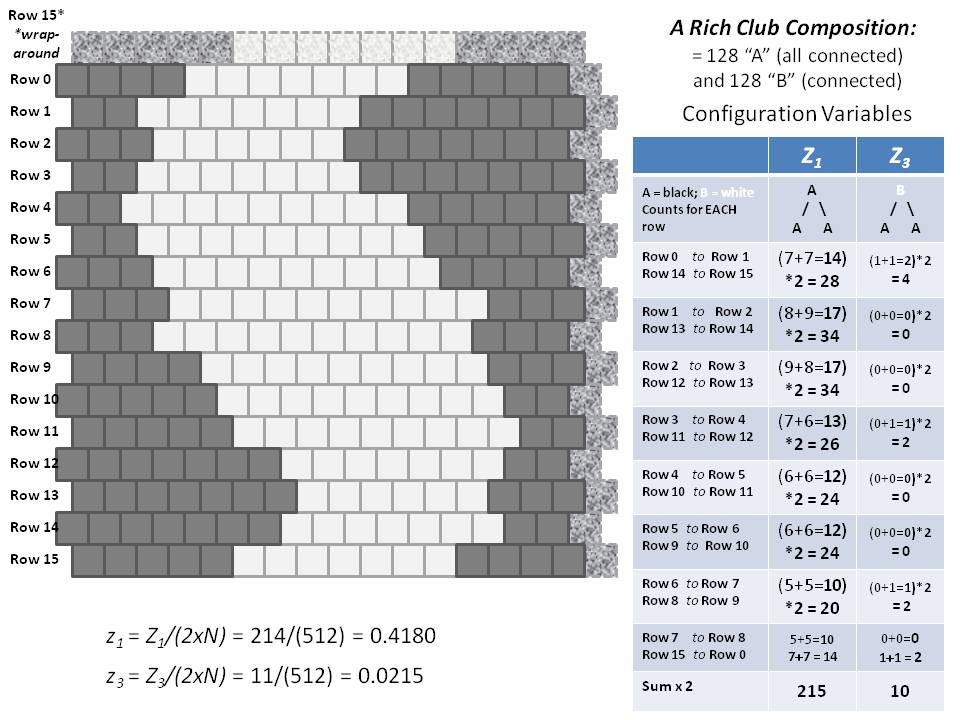}} 
  \vspace{3mm} 
  \caption{A 2-D CVM ``extreme-rich-club-like'' system with an equal number state \textbf{A} and state \textbf{B} nodes (128 nodes each).}   
\label{fig:CVM-2D_rich-club_cleaned-up_v2_2017-12-12}
\end{figure}

\textbf{\textit{Note:}} The exact details of the row counts are difficult to read in Figures~\ref{fig:CVM-2D_scale-free_2017-12-12} and~\ref{fig:CVM-2D_rich-club_cleaned-up_v2_2017-12-12}; the original diagrams are in a corresponding slidedeck that will be available in the associated GitHub repository; see details at the end of this document. 

\textbf{\textit{Note:}} The count for the $z_i$ variables is approximate; not exact. A follow-on code analysis revealed that -- while the counting steps for the $x_i$, $y_i$, and $w_i$ configuration variables was precise, the counting for the $z_i$ configuration variables was done only across the horizontally-expressed variables, and did not include the vertical versions. This was true for both the computer code and the manual counting. Because the size and diversity of patterns within any of the testing grids was sufficient to give a reasonably accurate result for the $z_i$, I decided to keep the code as is. This was verified via manual counts on some very small-scale 2-D grids. Another reason to stay with the current code (for approximate results) is that the next step will be a transiston to a full object-oriented approach, and that the time spent on code revision would be best served by moving on to the next stage. 

The second configuration, for an ``extreme-rich-club-like'' configuration, is shown in Figure~\ref{fig:CVM-2D_rich-club_cleaned-up_v2_2017-12-12}.

\vspace{3mm} 

As a contrast to the grid configuration used in Figures~\ref{fig:CVM-2D_Scale-free_128-nodes_pg0_2016-10-26} and~\ref{fig:CVM-2D_scale-free_2017-12-12}, I created a second configuration that had only one large compact region of nodes in state \textbf{A}, which was wrapped-around the grid envelope, as shown in Figure~\ref{fig:CVM-2D_rich-club_cleaned-up_v2_2017-12-12}. This configuration was designed to maximize the number of pairwise and triplet configurations that put ``like-near-like.'' The previous configuration, shown in Figure~\ref{fig:CVM-2D_Scale-free_128-nodes_pg0_2016-10-26}, was more in the direction of ``like-near-unlike.'' 

The purpose of having configurations with such different dispersions among the configuration variable values was that they would putatively yield different \textit{h}-values, or correspond to different points on an equilibrium curve for the free energy equation (in the case of equiprobable units in states \textbf{A} and \textbf{B}). As I have analytic results for that free energy minimum curve (the equilibrium point for the free energy at different \textit{h}-values, or interaction enthalpy values), it would serve as both a useful experiment and V\&V test. These results are discussed in the following Section~\ref{sec:v-and-v-thermodynamic-variables}. 

The V\&V for the initial stage of code development; ascertaining that the configuration variable counts were as they should be, was complete. 

There is an accompanying slidedeck that documents the code block structure and provides other important elements of code documentation (other than V\&V); this will also be available on GitHub; see the end of this document for details. 

By far, the most complex element of the ``configuration variable counting'' code was in counting the triplets. The V\&V step ensured that the counts wrapping around from right to left, and from top to bottom (creating a completely-wrapped envelope of the initial 2-D grid) performed as desired and expected. (See the \textit{Note} earlier in this section; the counts for the $z_i$ variables are done in the horizontal direction only, for both the code and manual verification.)

\section{Verification and Validation of Computing the Thermodynamic Variables}
\label{sec:v-and-v-thermodynamic-variables}
%

There were two primary means for obtaining validation that the code computing the thermodynamic variables was correct: 

\begin{enumerate} \itemsep0pt 
\item \textbf{Comparison with analytic for the equiprobable case} -- for the case equiprobable distribution among the $x_i$ variables ($x_1 = x_2 = 0.5$), I have developed an analytic solution, which gives a means for comparing the code-generated results against the expected (analytic) results, and 
\item \textbf{Comparison with analytic for the case where the interaction enthalpy is zero} -- the second means to check the code-generated results is for the case where the distributuion of $x$ values is not equiprobable, however the interaction enthalpy is set to zero ($h = 1$), and thus the exact distribution of other configuration values can be precisely computed, allowing further for exact analytic computation of thermodynamic variables. 
\end{enumerate}

The previous section described the patterns generated for the validation of configuration variable counting. It was interesting to see how the thermodynamic variables emerged for the systems described there, however (as will be illustrated here), certain of those system were \textit{not} at equilibrium, even though they had equiprobable distribution of $x_i$ values. As these results are more in the realm of theory and less V\&V, they will be discussed elsewhere. 

The realization that manually-generated patterns would \textit{not} necessarily be at equilibrium meant that I needed to have test cases where the patterns \textit{would indeed be at equilibrium}; this required not only random generation of patterns, but also that they be modified so that their associated free energy values achieved minimum. This generation-and-modification process is described more thoroughly in the following Section~\ref{sec:v-and-v-free-energy-minimization}. 

\subsection{Validation support: analytic solution}
\label{subsec:validation-support-analytic-solution}

The analytic solution for the case where $x_1 = x_2 = 0.5$ can be found when we are using the full interaction enthalpy term of $\varepsilon_1*(2y_2-y1-y3)$. This solution is similar to the more limited enthalpy equation, used in \cite{Maren_2016_CVM-primer-neurosci} as well as in the predecessor work \cite{AJMaren-TR2014-003}, where $\varepsilon_1*(2y_2)$.

The free energy equation for a 2-D CVM system, including configuration variables in the entropy term, is

\begin{equation}
\label{Bar-G-2-D-basic-eqn}
  \begin{aligned}
\bar{G}_{2-D} = G_{2-D}/N = \\
  & \varepsilon(z_2+z_3+z_4+z_5) \\
- & \Bigg[ 2 \sum\limits_{i=1}^3 \beta_i Lf(y_i)) 
          + \sum\limits_{i=1}^3 \beta_i Lf(w_i)) 
          - \sum\limits_{i=1}^2 \beta_i Lf(x_i)) 
          - 2 \sum\limits_{i=1}^6 \gamma_i Lf(z_i) \Bigg]\\
+ & \mu (1-\sum\limits_{i=1}^6 \gamma_i  z_i )+4 
\lambda (z_3+z_5-z_2-z_4)
  \end{aligned}
\end{equation}

\noindent
where $\mu$ and $\lambda$ are Lagrange multipliers, and  we have set $k_{\beta}T = 1$. 

\textit{Note:} the full derivation of the 2-D CVM free energy is presented in \cite{AJMaren-TR2014-003}, and the preceding equation corresponds to  Equations (2)--(14) of that reference. However, that derivation was for the case where the interaction enthalpy used $2y_2$ and not $2y_2 - y_1 - y_3$; the methods for the two derivations are similar, and the results are scaled relative to each other. 

Also, the single enthalpy parameter here is $\varepsilon$, with the enthalpy parameter for unit activation implicitly set to zero, as the earlier intention was to solve the above equation for an analytic solution, which was possible only in the case where $x_1 = x_2 = 0.5$, meaning that the per-unit enthalpy activation parameter $\varepsilon_0 = 0$.

The enthalpy term used previously, in \cite{Maren_2016_CVM-primer-neurosci} and in \cite{AJMaren-TR2014-003}, was

\begin{equation}
\label{Eqn:z3-analyt2-previous-approach}
  \begin{aligned}
\bar{H}_{2-D} = H_{2-D}/N =
\varepsilon_1(2y_2) = \varepsilon_1(z_2+z_3+z_4+z_5). 
  \end{aligned}
\end{equation}

The approach that I am using currently is to take the same enthalpy equation as originally advocated by Kikuchi \cite{Kikuchi_1951_Theory-coop-phenomena} and by Kikuchi and Brush \cite{Kikuchi-Brush_1967_Improv-CVM}, which gives 

\begin{equation}
\label{Eqn:z3-analyt1-current-approach}
  \begin{aligned}
\bar{H}_{2-D} = H_{2-D}/N = 
\varepsilon_1(2y_2 - y_1 - y_3) =
\varepsilon_1(z_4+z_3-z_1-z_6). 
  \end{aligned}
\end{equation}

Both of these equations are found using equivalence relations, specifically

\begin{eqnarray}
\label{Eqn:equivalence-relationships-y-and-z}
  y_2 = z_2+z_4 = z_3+z_5 \\
  2 y_2 = z_2+z_4 + z_3+z_5.   
\end{eqnarray}

Kikuchi and Brush \cite{Kikuchi-Brush_1967_Improv-CVM} found an analytic solution for this equation, for the condition where $x_1 = x_2 = 0.5$. They presented their solution, without derivation, in their 1967 paper. I re-derived the same analytic solution, and presented it (albeit for the case where the interaction enthalpy used $2y_2$ and not $2y_2 - y_1 - y_3$) in \cite{AJMaren-TR2014-003}. (Maren \cite{Maren_2016_CVM-primer-neurosci} gave the details for the analytic solution for the 1-D CVM; the derivation for the 2-D CVM is similar.)

The solution for the case where the enthalpy term involves only $\varepsilon_1 2y_2$, for the condition where $x_1 = x_2 = 0.5$, is given as

\begin{equation}
\label{Eqn:z3-analyt2-previous-approach-solution}
  z_3=\frac{(h^2-3)(h^2+1)}{8[h^4-6h^2+1]}. 
\end{equation}\\
\vspace{-12pt}

When the more complete enthalpy expression is used, viz. $\varepsilon_1(2y_2-y_1-y_3) = \varepsilon_1(z_4+z_3-z_1-z_6)$, the analytic solution becomes 

\begin{equation}
\label{Eqn:z3-analyt1-current-approach-solution}
  z_3=\frac{(h^4-3)(h^4+1)}{8[h^8-6h^4+1]}. 
\end{equation}\\
\vspace{-12pt}

(\textit{Note:} the full derivation of these results will be published separately.)

The experimentally-generated results from probabilistically-generated data sets correspond to the analytic results in the neighborhood of $h=0$. The reason that the range is so limited is that the analytic solution makes use of equivalence relations as expressed above. 

The resulting solution has divergences at $h^4 = 0.172$ and $h^4 = 5.828$, corresponding to divergences when  $h = 0.644$ and  $h = 1.554$. We are interested in the latter case, where the value of $h>1$ indicates that $\varepsilon1 >0$, which is the case where the interaction enthalpy favors like-near-like interactions, or some degree of gathering of similar units into clusters. This means that we expect that the computational results would differ from the analytic as $h \rightarrow 1.554$. 

The comparison is shown in the following Figure~\ref{fig:Perturbation-Results_x-eq-0pt50_analytic_crppd_2018-01-22}. In this figure, the column in the table marked as \textbf{z3Analyt1} corresponds to results from Eqns.~\ref{Eqn:z3-analyt1-current-approach} and \ref{Eqn:z3-analyt1-current-approach-solution} (the current approach) and in the next column, \textbf{z3Analyt2} corresponds to results from Eqns.~\ref{Eqn:z3-analyt2-previous-approach} and \ref{Eqn:z3-analyt2-previous-approach-solution} (the previous approach).

\begin{figure}[ht]
  \centering
  \fbox{
  \rule[-.5cm]{0cm}{4cm}\rule[-.5cm]{0cm}{0cm}	
  \includegraphics [trim=0.0cm 0cm 0.0cm 0cm, clip=true,   width=0.95\linewidth]{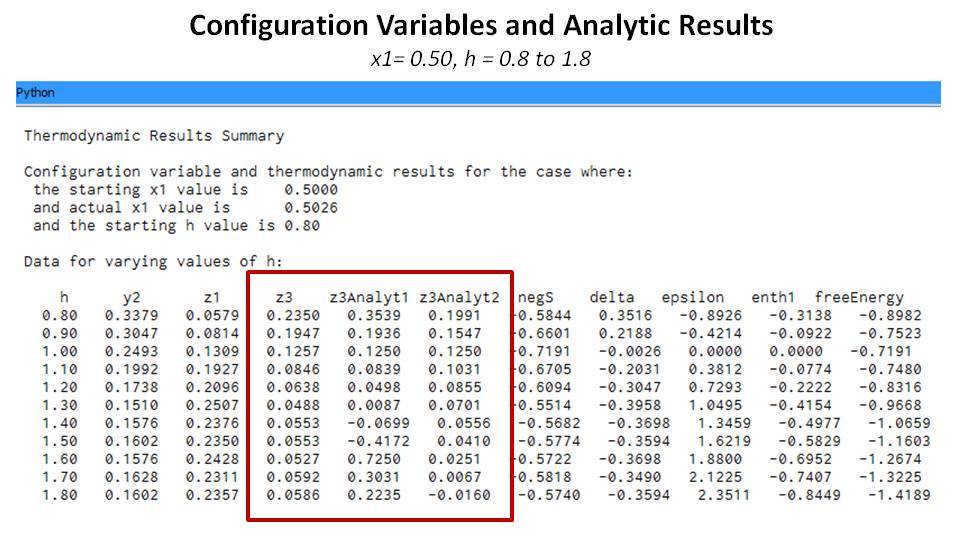}} 
  \vspace{3mm} 
  \caption{Data table giving the probabilistically-generated $z_3$ results (after reaching free energy minimum) vs. the analytic results for two different formulations of the enthalpy expression, all for the case where $x_1 = x_2 = 0.5$, and where $h = 0.8 .. 1.8$.}   
\label{fig:Perturbation-Results_x-eq-0pt50_analytic_crppd_2018-01-22}
\end{figure}
\vspace{3mm} 

The graph is shown in the following Figure~\ref{fig:Perturbation-Results_x-eq-0pt50_analytic_graph_2018-01-22}.

\begin{figure}[ht]
  \centering
  \fbox{
  \rule[-.5cm]{0cm}{4cm}\rule[-.5cm]{0cm}{0cm}	
  \includegraphics [trim=0.0cm 0cm 0.0cm 0cm, clip=true,   width=0.90\linewidth]{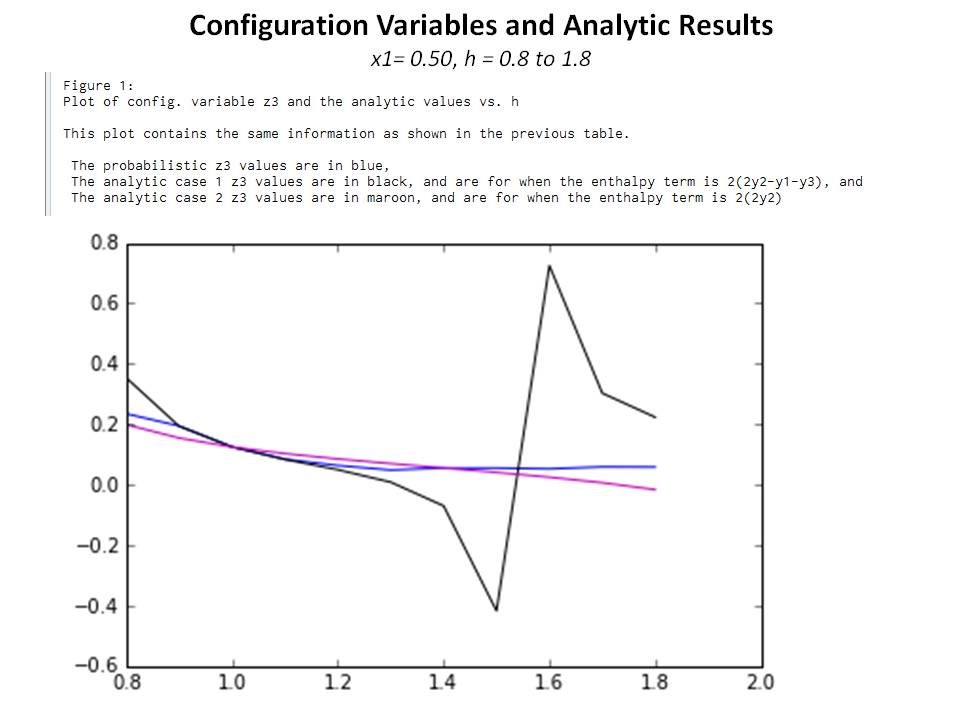}} 
  \vspace{3mm} 
  \caption{Graph giving the probabilistically-generated $z_3$ results (after reaching free energy minimum) vs. the analytic results for two different formulations of the enthalpy expression, all for the case where $x_1 = x_2 = 0.5$, and where $h = 0.8 .. 1.8$. As discussed in the body of this work, the analytic solution diverges at $h = 1.554$, where the denominator becomes zero.}   
\label{fig:Perturbation-Results_x-eq-0pt50_analytic_graph_2018-01-22}
\end{figure}
\vspace{3mm} 

The divergent behavior in the analytic solution is likely due to the use of equivalence relationships, as identified in Eqn.~\ref{Eqn:equivalence-relationships-y-and-z}.

%
\subsection{Validation support: basic thermodynamic results}
\label{subsec:validation-support-thermodynamics-basic}
%

The following Figure~\ref{fig:Perturbation-Results_x1=0pt50_lbld_crppd_2018-01-22} shows the results when $x_1 = 0.5$, which is the case where all of the results should conform with the analytic solution.

\begin{figure}[ht]
  \centering
  \fbox{
  \rule[-.5cm]{0cm}{4cm}\rule[-.5cm]{0cm}{0cm}	
  \includegraphics [trim=0.0cm 0cm 0.0cm 0cm, clip=true,   width=0.95\linewidth]{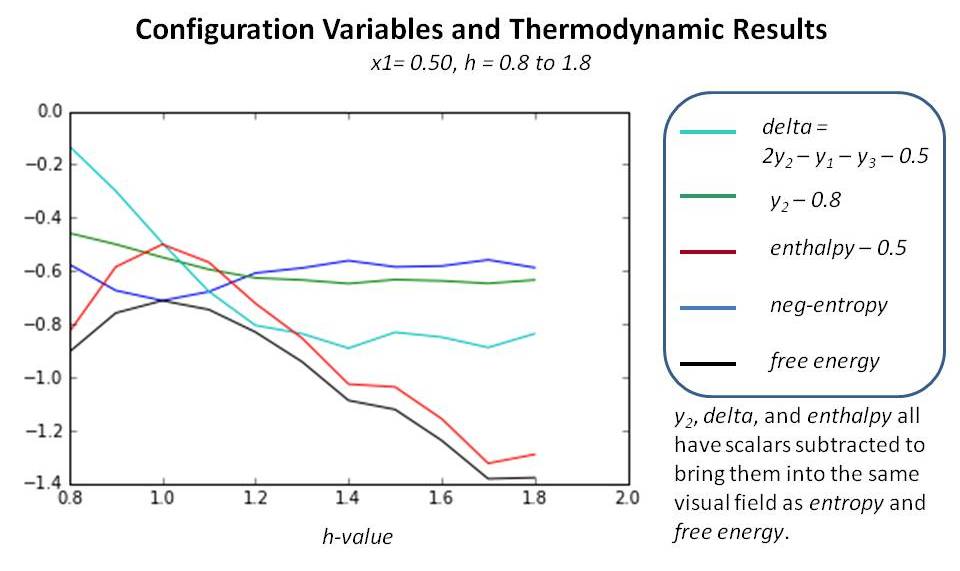}} 
  \vspace{3mm} 
  \caption{Configuration variable and thermodynamic values for the case where $x_1 = x_2 = 0.5$, and where the interaction enthalpy parameter $h$ ranges as $h = 0.8 .. 1.8$. See detailed explanation of results in the following Section~\ref{sec:v-and-v-free-energy-minimization}, as their nature is similar to these results. }   
\label{fig:Perturbation-Results_x1=0pt50_lbld_crppd_2018-01-22}
\end{figure}
\vspace{3mm} 

\begin{figure}[ht]
  \centering
  \fbox{
  \rule[-.5cm]{0cm}{4cm}\rule[-.5cm]{0cm}{0cm}	
  \includegraphics [trim=0.0cm 0cm 0.0cm 0cm, clip=true,   width=0.95\linewidth]{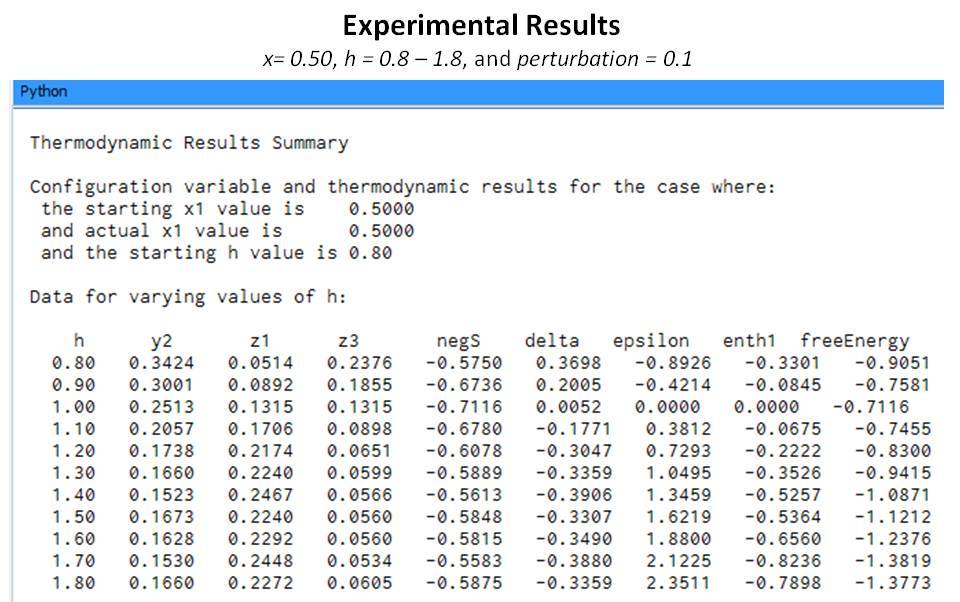}} 
  \vspace{3mm} 
  \caption{Data table for configuration variable and thermodynamic values for the case where $x_1 = x_2 = 0.5$, and where the interaction enthalpy parameter $h$ ranges as $h = 0.8 .. 1.8$. }   
\label{fig:Perturbation-Results_x1=0pt50_data-table_crppd_2018-01-22}
\end{figure}

The corresponding Figure~\ref{fig:Perturbation-Results_x1=0pt50_data-table_crppd_2018-01-22} presents the data table supporting Figure~\ref{fig:Perturbation-Results_x1=0pt50_lbld_crppd_2018-01-22}.

\vspace{3mm} 

\section{Verification and Validation of Free Energy Minimization}
\label{sec:v-and-v-free-energy-minimization}
%

It is not enough to simply compute the thermodynamic variables for a given 2-D grid configuration; it is important to have a mechanism by which the pattern of node activations on the grid can adjust in order to reach a free energy minimum. 

I accomplished this by writing the code for two stages: 

\begin{enumerate} \itemsep0pt 
\item \textbf{\textit{Bring $x_1$ close to desired value}}, and
\item \textbf{\textit{Adjust configuration variables to achieve free energy minimum}}.
\end{enumerate} 

\textbf{\textit{Adjusting total number of nodes to achieve desired $x_1$:}}

The code has an initial specification for the desired $x_1$ value (in **main**), and randomly generates a 2-D CVM grid according to a probabilistic assignment of ``1'' (state \textbf{A}) or ``0'' (state \textbf{B}) to the units in the grid. However, just because the probability (of the random number generation) is set to a specified value (say, 0.35) does not mean that the resulting total of state A nodes will be precisely 0.35 of the total number of nodes (e.g., 0.35 * 256, or 90 nodes); thus, a few nodes will have to be ``flipped'' in order to bring the actual number of nodes in state \textbf{A} closer to the desired value. 

The code specifies a tolerance value for how close the actual $x_1$ value needs to be to the desired value. It runs a function to randomly select and flip unit values (as needed, going in the right direction), and continues this until the resulting actual $x_1$ is within desired tolerance. 

\textbf{\textit{Validation: Printing out the actual values for $x_1$, ensuring that they are within tolerance of the desired value. }}

\vspace{10 pt}
 
\textbf{\textit{Adjusting configuration variables to achieve free energy minimum:}}

There is no guarantee (in the current version of the code) that the free energy minimum is actually met; instead, the code will run this entire process (generating a new grid, adjusting for $x_1$ within tolerance, and then adjusting the units so that free energy is progressively decreased) for a specified number of trials. During the debug phase, the number of trials was between 1 - 3, so that I could closely monitor the process. During actual runs, the trials were typically 10 - 20; there was not much variability in the results. 

The goal of this process is to keep adjusting the grid units so that free energy is decreased. For each run, there is a constant value for $x_1$. That means, before any nodes are flipped, the program will (randomly) find a node in state \textbf{A}, and another node in state \textbf{B}. It will flip the two (from state \textbf{A} to state \textbf{B}, and vice versa). It will then compute the new free energy; this requires recomputing the entire set of configuration variable values. While $x_1$ is held constant with this process, it is likely that all other configuration variables ($y_i$, $w_i$, and $z_i$) will change. 

The program computes the new free energy value (using the new configuration variable values as well as the \textit{h}-value that is being tested for the run). If the free energy is lower, the change in the units is kept. If not, both units are reverted back to their original values. 

The trials are strictly probabilistic for this generation of code development; there is no attempt to find nodes whose topographic position (i.e., sets of neighbors, nearest-neighbors, and triplets) would be most likely to produce a free energy decrease if the node were to change. 

One version of the code is designed less to run multiple trials, and more to collect, print, and plot the thermodynamic variables over a series of attempts to flip nodes and test the resulting free energy. 

One validation step is visual observation of the thermodynamic variables over the course of any one of these trials; noting that the free energy does, in fact, decrease. 

Another validation step is that when $h=1$ ($\varepsilon_1 = 0$), there is no interaction energy. In this case, the final configuration variable values should be very close to their probabilistic likelihoods. Thus, for example, when $h=1$ and $x_1 = 0.35$, we expect that $y_1 = 0.35*0.35 = 0.1225$, etc. Thus, it is possible to compare the actual resultant configuration variable values with the probabilistic expectancies. 

A final validation step is to compare the resulting behaviors against the theoretical expectations. This is discussed more fully in the following subsection.

\subsection{Validation Support: Exemplar Code Run}
\label{subsec:validation-support-code-run}

An example is shown in the following Figure~\ref{fig:Perturbation-Results_x1=0pt35_lbld_crppd_2018-01-22}. 

This data is actually from a perturbation run, where the 2-D grid is established as previously described, and then perturbed by a given amount (in this case, a fraction of 0.1 of the existing nodes are flipped), and then taken to free energy minimum a second time.

\begin{figure}[ht]
  \centering
  \fbox{
  \rule[-.5cm]{0cm}{4cm}\rule[-.5cm]{0cm}{0cm}	
  \includegraphics [trim=0.0cm 0cm 0.0cm 0cm, clip=true,   width=0.95\linewidth]{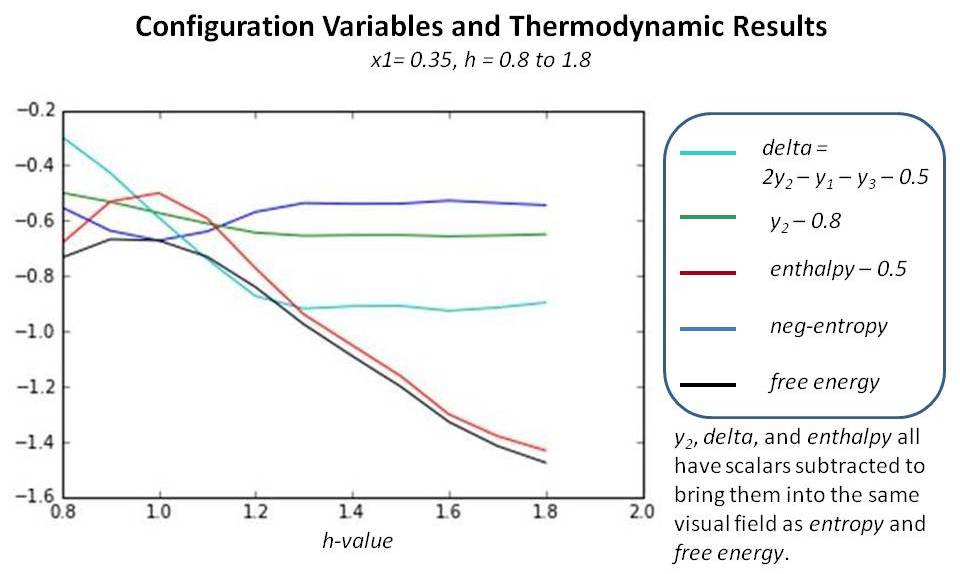}} 
  \vspace{3mm} 
  \caption{Configuration variable and thermodynamic values for the case where $x_1 = 0.35$ and $x_2 = 0.65$, and where the interaction enthalpy parameter $h$ ranges as $h = 0.8 .. 1.8$. See detailed explanation of results within Section~\ref{sec:v-and-v-free-energy-minimization}.}   
\label{fig:Perturbation-Results_x1=0pt35_lbld_crppd_2018-01-22}
\end{figure}
\vspace{3mm} 

These results were obtained from the program 2D-CVM-perturb-expt-1-2b-2018-01-12.py, run on Friday, Jan. 12, 2018.

\begin{figure}[ht]
  \centering
  \fbox{
  \rule[-.5cm]{0cm}{4cm}\rule[-.5cm]{0cm}{0cm}	
  \includegraphics [trim=0.0cm 0cm 0.0cm 0cm, clip=true,   width=0.95\linewidth]{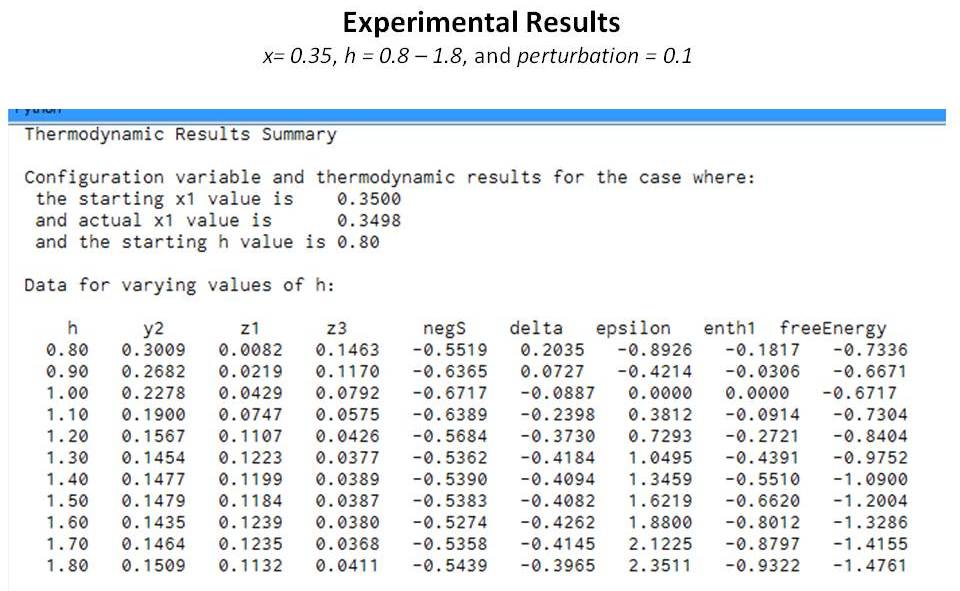}} 
  \vspace{3mm} 
  \caption{Data table containing configuration variable and thermodynamic values for the case where $x_1 = 0.35$ and $x_2 = 0.65$, and where the interaction enthalpy parameter $h$ ranges as $h = 0.8 .. 1.8$. See detailed explanation of results within Section~\ref{sec:v-and-v-free-energy-minimization}.}   
\label{fig:Perturbation-Results_x1=0pt35_data-table_crppd_2018-01-18}
\end{figure}

The parameter settings were for $x_1 = 0.35$ and $h = 0.8 .. 1.8$, with a total of twenty trials ($numTrials = 20$) for each \textit{h}-value.  A data table from this run is shown in Figure~\ref{fig:Perturbation-Results_x1=0pt35_data-table_crppd_2018-01-18}. All reported results for configuration variable and thermodynamic values are averages over $numTrials$ runs, where $numTrials = 20$.

\subsubsection{Validation support: $y_2$ results}
\label{subsubsec:validation-support-y2}

The values observed for $y_2$ conform to expectations. In Figure~\ref{fig:Perturbation-Results_x1=0pt35_lbld_crppd_2018-01-22}, \textbf{$y_2$} is shown in green, as $y_2 - 0.8$ (in order to bring the $y_2$ values within the same visual range as other results). 

When $h=1.0$, $y_2 = 0.2278$, which is the expected result. (The true expected results is $y_2 = 0.35 * 0.65 = 0.2275$; the observed value of $0.2278$ is an average over twenty trials. The deviance from the theoretical expectation is acceptable. )

\vspace{3mm} 

When $h<1.0$, the $y_2$ values are greater, and when $h>1.0$, the $y_2$ values are smaller. In fact, $y_2$ ranges from $y_2 = 0.301$ (when $h = 0.8$) down to $y_2 = 0.151$ (when $h = 1.8$). These again are expected results. A separate document will address the theoretical expectations in more detail. In brief, when $h<1.0$, then $\varepsilon_1 <0.0$, meaning that the interaction enthalpy parameter $\varepsilon_1$ is negative. When $\varepsilon_1$ is negative, the enthalpy is decreased by increasing $y_2$, as the interaction enthalpy $\varepsilon_1$ multiplies the term $(2 y_2 - y_1 - y_3)$. Thus, maximizing $y_2$ is expected when $h<1.0$.

There is a limit as to how far $y_2$ can be increased; presumably it can approach $0.5$, however, that would mean that the units were arranged in a strict checkerboard manner; that there were no instances of like-near-like at all. This is rather difficult to achieve; both in creation of highly-ordered systems, and in this particular code, which uses a simplistic \textit{find-and-flip} strategy. 

As previously noted, when $h>1.0$, the $y_2$ values are smaller. This is the case where $\varepsilon_1 >0.0$, and system enthalpy is decreased when $y_2$ is made smaller. Thus, the system moves more towards a like-with-like configuration (increasing $y_1$ and $y_3$); maximizing the size of the various ``islands,'' and decreasing the size of their borders (minimizing $y_2$). 

There is a practical limit as to how far $y_2$ can be decreased; there will always be a border area between the state \textbf{A} islands (or even a single, massive state \textbf{A} continent) and the surrounding sea of state \textbf{B} units. This means that $y_2$ will not get close to zero. The actual practical limit for $y_2$ will actually depend on the total system size (total number of nodes), because the border area will progressively decrease (although not disappear) as more and more islands join to become continents. Thus, the value of $y_2 < 0.157$, which occurs when $h \geq 1.2$, is not surprising. 

Once $y_2$ is pushed to a suitably small value, it becomes increasingly difficult for the simple \textit{find-and-flip} strategy to (randomly) find nodes where the flip will accomplish a free energy reduction. This is likely why there is general stability in the $y_2$ values beyond $h \geq 1.2$; there are simply not that many nodes where the flip will do much good, keeping in mind that \textit{two} nodes (each in a different state) have to be flipped in order to maintain the $x_1$ value.    

Thus, the preliminary conclusion is that free energy minimization is being accomplished, and that the $y_2$ values are behaving as expected.

\subsubsection{Validation support: $delta$ results}
\label{subsubsec:validation-support-delta}

Again referencing Figure~\ref{fig:Perturbation-Results_x1=0pt35_lbld_crppd_2018-01-22}, we examine the curve for \textit{delta} (shown in cyan), defined as $(2 y_2 - y_1 - y_3)$, which is the actual term that is multiplied by $\varepsilon_1$ to achieve the interaction enthalpy term. (The \textit{delta} curve is shown in dark green in this figure.) This curve behaves as expected. 

In particular, we note that there is a nearly linear behavior in the region between $h = 0.8$ and $h = 1.3$. When $h = 0.8$, \textit{delta} = 0.2035 (according to the data table shown in Figure~\ref{fig:Perturbation-Results_x1=0pt35_data-table_crppd_2018-01-18}. When $h = 1.2$, \textit{delta} = -0.3730. When $h = 1.0$, we would expect that there would be purely probabilistic distribution of units into their configurations, and thus expect that $y_1 = 0.35*0.35 = 0.1225$, $y_3 = 0.65*0.65 = 0.4225$, and $y_2 =.2275$ (as mentioned earlier). We would have then that $2 y_2 - y_1 - y_3 = 2*.2275 - 0.1225 - 0.4225 = -0.090$. The actual value is \textit{delta} = -0.0887, which is acceptably close. 

Similar arguments hold for the expected and observed values of \textit{delta} as did for $y_2$ in the preceding discussion. 

We again note that the values for \textit{delta} level out as \textit{h} increases beyond 1.2; this is because there are not that many units that the simple \textit{find-and-flip} strategy can easily find. In particular, we note that the $z_3$ value at $h \geq 1.3$ is typically around $z_3 = 0.04$, which is very small. 

In particular, we observe that this $z_3$ value indicates that we have pushed the system to its limit for minimizing $z_3$, which is the \textbf{A}-\textbf{A}-\textbf{B} configuration. This $z_3$ value indicates a border of a rather large island of state \textbf{A} units in a sea of \textbf{B} units. Specifically, for the 256-unit system that is the subject for this investigation, when $z_3 = 0.04$, then $N = 0.4*256/2 = 102.4/2 = 51$ triplets involve border units around islands / continents of state \textbf{A}. This is approximately \textit{1/5th} of the total number of units available. This suggests that we have pushed the system about as far as it can go. Of course, a visual inspection of the resulting grid would be enormously useful in confirming these assessments. This will be included in a subsequent document.  

\subsubsection{Validation support: thermodynamic results}
\label{subsubsec:validation-support-thermodynamics}

The enthalpy is maximum when $h=0$, which is to be expected. As we minimize free energy, we minimize enthalpy. As soon as we introduce some non-zero interaction enthalpy, we have an opportunity to adjust the configuration values (specifically the $y_i$, as just discussed) to lower the enthalpy. 

The entropy is similarly at a maximum (neg-entropy is at a minimum) when $h=0$. The negative entropy increases for non-zero values of \textit{h}, as expected. 

We particularly note that in the vicinity of $h=0$, or more generally, in the range of $0.8 \leq h \leq 1.3$, the variances in the entropy and enthalpy are approximately on the same scale; one does not appreciably dwarf the other. 

When we move beyond $h \geq 1.3$, we find that the enthalpy term strongly dominates the entropy, and thus dominates the free energy. It does this because we are increasing the value of the interaction enthalpy coefficient, $\varepsilon_1$, and not because we are gaining any appreciable difference in the configuration values. As noted in the previous discussions, these values have more-or-less stabilized in this range. 

Thus, increasing $h$ beyond $h = 1.3$ does not serve any useful value, suggesting a practical bound on \textit{h}-values for this kind of system. 

Our actual and practical choices for the \textit{h}-values should be based on the kind of behavior that we want to see in the configuration values. For modeling brain-like systems, we will most likely want $h \geq 0$, as that induces \textit{like-with-like} clustering, which seems to characterize certain neural collectives.

\section{Code included in this V\&V description}
\label{sec:code-included-in-V-and-V}
%

\begin{itemize}
\setlength{\itemsep}{1pt}
\item \textbf{2D-CVM-perturb-expt-1-2b-2018-01-12.py} - perturbation analysis with user-specifiable (in **main**) values for $x_1$, $h$, $numTrials$, and many other parameters. 
\end{itemize}

\textbf{Code Availability:} All code referenced here will be made available on a public GitHub repository after a short delay from initial publication of this V\&V document. This and related code will be supported by extensive documentation, which will also be placed in the GitHub repository. Anyone desiring access to the original code prior to its placement on a public GitHub repository should contact A.J. Maren at: alianna@aliannajmaren.com.

\textbf{Copyright:} All code referenced here has been independently develeoped by A.J. Maren. A.J. Maren holds the copyright to both the code and this V\&V document itself. arXiv is granted non-exclusive and irrevocable license to distribute this article.  

 \vspace{10 pt}


%
%

\end{document}